%% file: ms.tex
\documentclass[10pt,twocolumn,letterpaper]{article}
\pdfoutput=1
\usepackage{cvpr}              
\usepackage[accsupp]{axessibility}  

\usepackage{graphicx}
\usepackage{amsmath}
\usepackage{amssymb}
\usepackage{booktabs}
\usepackage{multirow}
\usepackage{bbding}
\usepackage{pifont}
\usepackage{MnSymbol}
\usepackage[dvipsnames]{xcolor}

\usepackage[pagebackref,breaklinks,colorlinks]{hyperref}

\usepackage{makecell}

\usepackage[capitalize]{cleveref}
\crefname{section}{Sec.}{Secs.}
\Crefname{section}{Section}{Sections}
\Crefname{table}{Table}{Tables}
\crefname{table}{Tab.}{Tabs.}
\usepackage[font=footnotesize,labelfont=bf]{caption}

\def\cvprPaperID{11663}
\def\confName{CVPR}
\def\confYear{2023}

\newcommand{\bd}{\mathbf{d}}

\newcommand{\bP}{\mathbf{P}}

\newcommand{\br}{\mathbf{r}}

\newcommand{\bW}{\mathbf{W}}

\newcommand{\bY}{\mathbf{Y}}
\newcommand{\bz}{\mathbf{z}}

\newcommand{\hl}{\hat{l}}

\newcommand{\hy}{\hat{y}}

\newcommand{\nR}{\mathbb{R}}

\newcommand{\cL}{\mathcal{L}}

\newcommand{\OurMethod}{GC-KPL}
\begin{document}

\title{3D Human Keypoints Estimation from Point Clouds in the Wild \\without Human Labels}

\author{Zhenzhen Weng$^{1}$\thanks{Work done as an intern at Waymo.} \quad Alexander S. Gorban$^2$\quad Jingwei Ji$^2$\quad Mahyar Najibi$^2$\quad \\
Yin Zhou$^2$\quad Dragomir Anguelov$^2$ \\ \\
$^1$Stanford University \quad $^2$Waymo 
}
\maketitle

\begin{abstract}
   Training a 3D human keypoint detector from point clouds in a supervised manner requires large volumes of high quality labels. While it is relatively easy to capture large amounts of human point clouds, annotating 3D keypoints is expensive, subjective, error prone and especially difficult for long-tail cases (pedestrians with rare poses, scooterists, etc.). In this work, we propose \OurMethod{} - Geometry Consistency inspired Key Point Leaning, an approach for learning 3D human joint locations from point clouds without human labels. We achieve this by our novel unsupervised loss formulations that account for the structure and movement of the human body. We show that by training on a large training set from Waymo Open Dataset \cite{sun2020scalability} without any human annotated keypoints, we are able to achieve reasonable performance as compared to the fully supervised approach. Further, the backbone benefits from the unsupervised training and is useful in downstream few-shot learning of keypoints, where fine-tuning on only 10 percent of the labeled training data gives comparable performance to fine-tuning on the entire set. We demonstrated that \OurMethod{} outperforms by a large margin over SoTA when trained on entire dataset and efficiently leverages large volumes of unlabeled data.
\vspace{-4mm}
\end{abstract}

\input{sections/introduction}
\input{sections/related_work}
\input{sections/method}

\input{sections/experiments}
\input{sections/conclusion}

\newpage
{\small
\bibliographystyle{ieee_fullname}

}

\end{document}


\title{Supplementary Material for\\
Unsupervised Learning of 3D Human Keypoints from Point Clouds in the Wild}

\author{Zhenzhen Weng$^{1}$\thanks{Work done as an intern at Waymo.} \quad Alexander S. Gorban$^2$\quad Jingwei Ji$^2$\quad Mahyar Najibi$^2$\quad \\
Yin Zhou$^2$\quad Dragomir Anguelov$^2$ \\ \\
$^1$Stanford University \quad $^2$Waymo 
}
\maketitle

\section{Synthetic Data Generation}
In our described Stage I, we initialize the model on a synthetic dataset that is constructed by ray casting onto randomly posed human mesh models (SMPL \cite{loper2015smpl}). Here we elaborate on the synthetic data generation process. We generate 1,000 16-frame sequences. Each sequence has a random SMPL body shape, and starts with the same standing pose and ends in a random pose. The poses in the middle of the sequence are linearly interpolated between the starting and ending poses. 

The ending pose was created by adding random noise to the rotation angles of each joint in the standing pose. To create realistic pedestrian poses, we add up to 60 degrees of random noise to the shoulder and elbow joint angles, and up to 30 degrees to the thigh and knee joints, and up to 5 degrees of noise to all other joints.

To simulate LiDAR point clouds, we place the human meshes at a distance of 6 to 17 meters from a ray caster and keep the faces that intersect with the rays. As in \cite{sun2020scalability}, we use 2650 vertical scans (with 360 degree coverage), and 64 LiDAR beams. We do not consider rolling shutter and other LiDAR artifacts for simplicity.

We construct 2-frame samples by taking consecutive frames from each sequence, and the same data augmentation is applied to both frames in each sample.

\section{Additional Qualitative Results}
\begin{figure}[t!]
    \centering
    \includegraphics[width=\columnwidth]{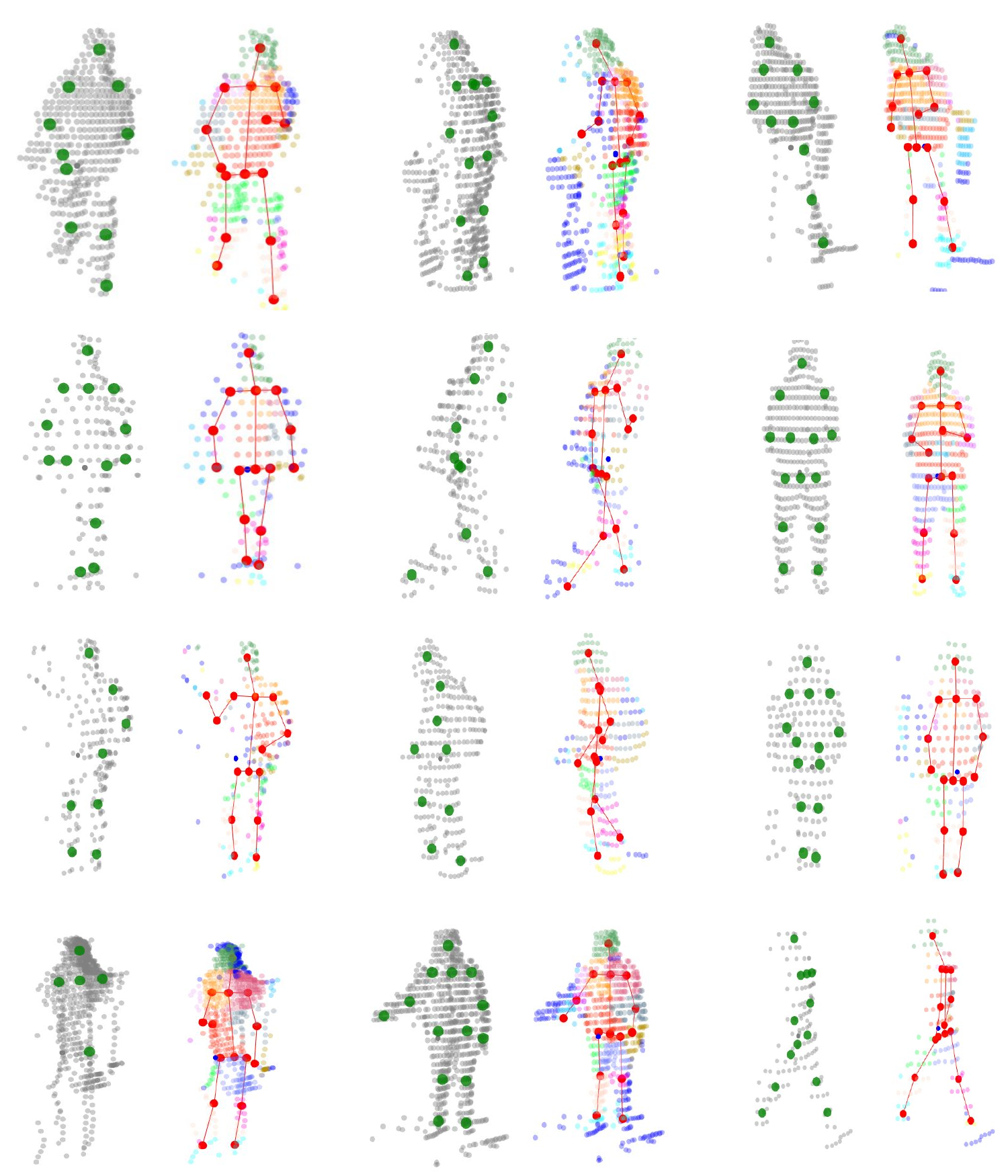}
    \caption{Additional qualitatve results. The points are colored by predicted segmentation labels. Ground truth keypoints are in \textcolor{OliveGreen}{green} and predicted keypoints and skeletons are in \textcolor{red}{red}.}
    \label{fig:additional_vis}
\end{figure}

\begin{figure}[h!]
    \centering
    \includegraphics[width=\columnwidth]{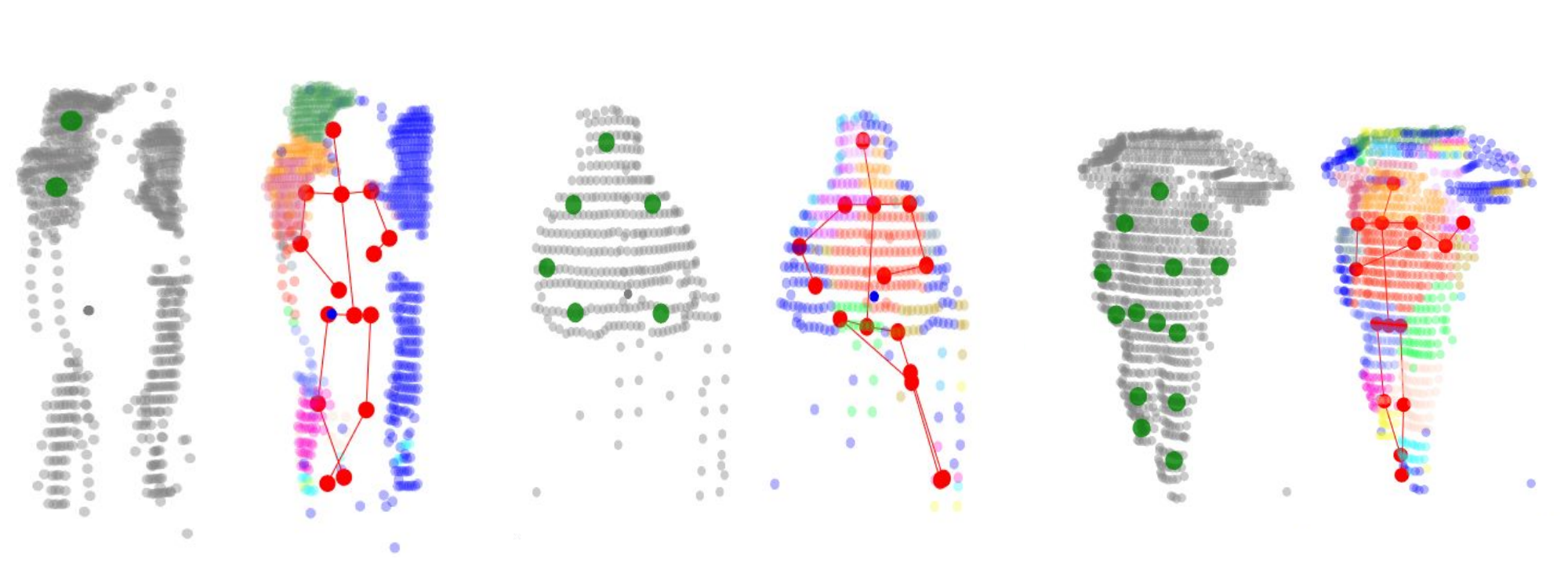}
    \caption{Failure cases.}
    \label{fig:failure_cases}
\end{figure}

In \cref{fig:additional_vis}, we include additional qualitative results from the finetuned (on 100$\%$ training data) model. We show typical failure cases on WOD in \cref{fig:failure_cases}, which are caused by occlusion (left and middle column) and incorrect segmentation of the point cloud (right column).

There is an animated visualization in the attachment. It demonstrates the effect of our unsupervised losses ($\cL_{flow}$, $\cL_{p2l}$ and $\cL_{sym}$). We perturb the ground truth keypoints by adding random noise (Gaussian noise with 0 mean and 6 cm standard deviation) to each keypoint. Then, we minimize these three losses with respect to the keypoints locations. We minimize with Adam optimizer with learning rate 1e-3 for 100 iterations. The weights for loss terms are $\lambda_{flow}=0.2$, $\lambda_{p2l}=0.1$, $\lambda_{sym}=5$. As shown, as the result of the optimization process the keypoints move to unperturbed locations over time.

{\small
\bibliographystyle{ieee_fullname}

}

%% file: sections/introduction.tex
\section{Introduction}
\label{sec:intro}

\begin{figure}
\centering
\includegraphics[width=\columnwidth]{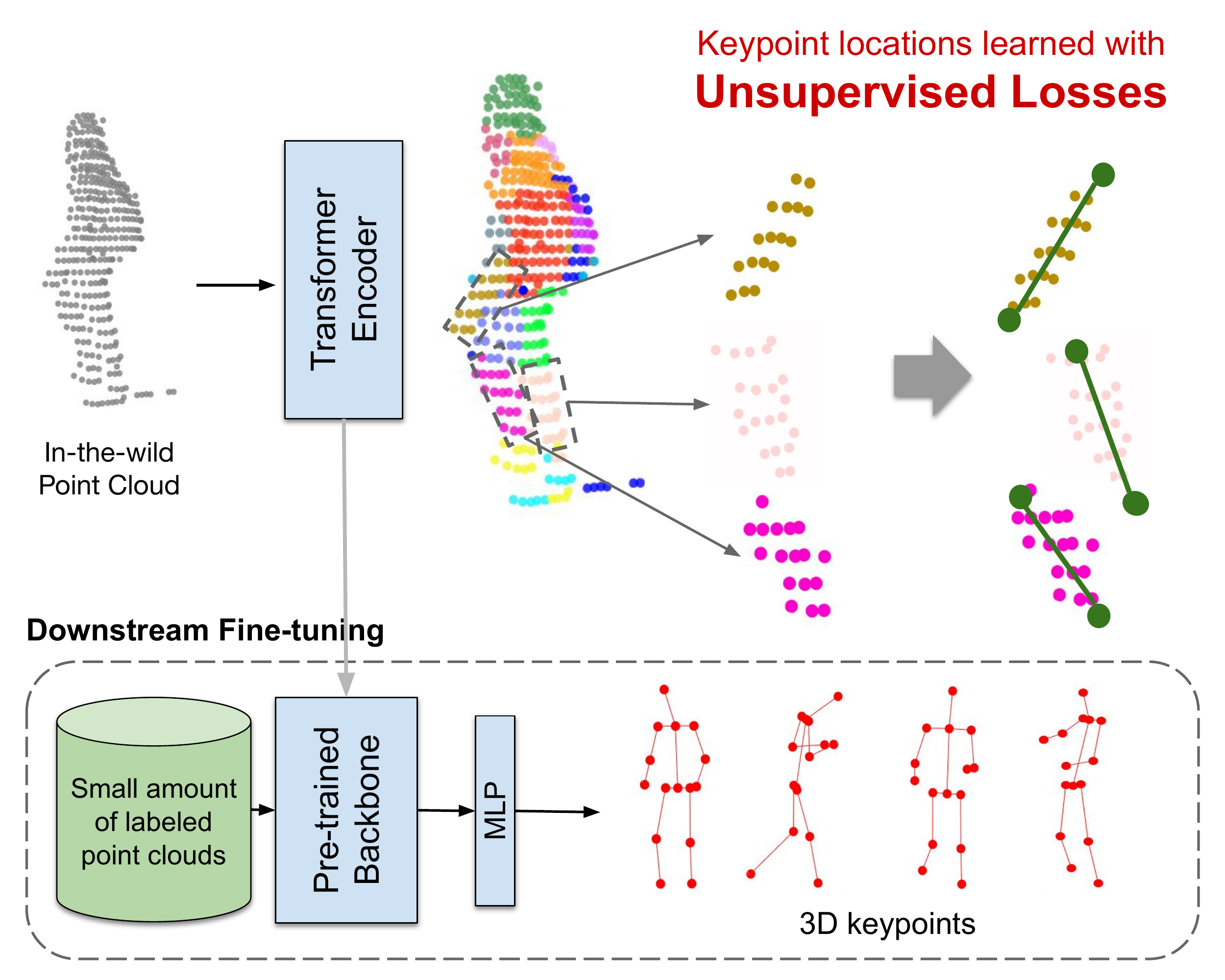}
\caption{We present \OurMethod, a novel method for learning 3D human keypoints from in-the-wild point clouds without any human labels. We propose to learn keypoint locations using unsupervised losses that account for the structure and movement of the human body. The backbone learns useful semantics from unsupervised learning and can be used in downstream fine-tuning tasks to boost the performance of 3D keypoint estimation.}
\label{fig:teaser}
\vspace{-4mm}
\end{figure}

Estimation of human pose in 3D is an important problem in computer vision and it has a wide range of applications including AR/VR, AI-assisted healthcare, and autonomous driving \cite{furst2021hperl,zanfirhum3dil,zheng2022multi}. For autonomous systems, being able to perceive human poses from sensor data (\eg LiDAR point clouds) is particularly essential to reason about the surrounding environment and make safe maneuvers.

Despite the high level of interest in human pose estimation in the wild, only few papers approached outdoor 3D keypoint detection using point cloud. A main reason is that training a pedestrian pose estimation model requires large amount of high quality in-the-wild data with ground truth labels. Annotating 3D human keypoints on point cloud data is expensive, time consuming and error prone. Although there are a few existing point cloud datasets with ground truth human poses \cite{sun2020scalability,li2022lidarcap,kim2019pedx}, they are limited in terms of the quantity of the 3D annotations and diversity of the data. Therefore, fully-supervised human keypoint detectors trained on such datasets do not generalize well for long tail cases. For this reason, previous approaches on pedestrian 3D keypoint estimation have mainly focused on utilizing 2D weak supervision \cite{zheng2022multi, furst2021hperl} which is easier to obtain, or leveraging signals from others modalities (\eg RGB, depth) \cite{zanfirhum3dil}. Nonetheless, there is a lot of useful information in the large amount of unlabeled LiDAR data that previous works on human pose estimation have not made an effort to utilize. 

In this work, we propose a novel and effective method for learning 3D human keypoints from in-the-wild point clouds without using any manual labeled 3D keypoints. Our approach is built on top of the key observation that human skeletons are roughly centered within approximately rigid body parts and that the location and movement of the surface points should explain the movement of the skeleton and vice versa. To that end, we design novel unsupervised loss terms for learning locations of the 3D keypoints/skeleton within human point clouds which correspond to 3D locations of major joints of human body.

In the proposed method, we first train a transformer-based regression model for predicting keypoints and a semantic segmentation model for localizing body parts on a synthetic data constructed from randomly posed SMPL human body model \cite{loper2015smpl}. Then, we train on the entire Waymo Open Dataset \cite{sun2020scalability} without using any 3D ground-truth annotation of human keypoints. Through unsupervised training, keypoint predictions are refined and the backbone learns useful information from large amount of unannotated data.

In summary, we make the following contributions:
\begin{itemize}
\item We present \OurMethod, a method for learning human 3D keypoints for in-the-wild point clouds without any manual keypoint annotations.
\item Drawing insight from the structure and movement of the human body, we propose three effective and novel unsupervised losses for refining keypoints. We show that the proposed losses are effective for unsupervised keypoint learning on Waymo Open Dataset.
\item Through downstream fine-tuning/few-shot experiments, we demonstrate that \OurMethod{} can be used as unsupervised representation learning for human point clouds, which opens up the possibility to utilize a practically infinite amounts of sensor data to improve human pose understanding in autonomous driving.
\end{itemize}

%% file: sections/related_work.tex
\section{Related Work}
\label{sec:related_work}

\subsection{3D Human Keypoint Estimation from Points Clouds}
There have been a few works \cite{shotton2011real,zhou2020learning,zhang2020weakly} about estimating 3D keypoints from clean and carefully-curated point clouds \cite{haque2016towards}, but 3D keypoint estimation from in-the-wild point clouds is a much less studied problem.
Due to the lack of ground-truth 3D human pose annotations paired with LiDAR data, there has not been a lot of works on 3d human keypoint estimation from LiDAR information. Among the few point cloud datasets with 3D keypoint annotations, LiDARHuman26M \cite{li2022lidarcap} captures long-range human motions with ground truth motion acquired by the IMU system and pose information derived from SMPL models fitted into point clouds. It is among the first few datasets which have LiDAR point clouds synchronized with RGB images, but SMPL shape parameters are same for all 13 subjects and it does not feature in-the-wild pedestrians where there could be much more background noise and occlusion. PedX \cite{kim2019pedx} offers 3D automatic pedestrian annotations obtained using model fitting on different modalities, gathered effectively from a single intersection with only 75 pedestrians (the second intersection has only 218 frames, labels for the third scene were not released). Waymo Open Dataset \cite{sun2020scalability} has more than 3,500 subjects from over 1,000 different in-the-wild scenes with high-quality 2D and 3D manual annotations. Despite the existence of these datasets, the few works on 3D pose estimation from point clouds mostly rely on weak supervision. HPERL model \cite{furst2021hperl} trains on 2D ground-truth pose annotations and uses a reprojection loss for the 3D pose regression task. Multi-modal model in \cite{zheng2022multi} uses 2D labels on RGB images as weak supervision, and creates pseudo ground-truth 3D joint positions from the projection of annotated 2D joints. HUM3DIL \cite{zanfirhum3dil} leverages RGB information with LiDAR points, by computing pixel-aligned multi-modal features with the 3D positions of the LiDAR signal. In contrast, our method does not use any RGB information or weak supervision.

\subsection{Unsupervised Keypoint Localization}
There are a number of works that aim to recover 3D keypoints using self-supervised geometric reasoning \cite{suwajanakorn2018discovery,li2019usip}, but they are limited to rigid objects. More recent unsupervised methods work for articulated objects from monocular RGB data \cite{jakab2018unsupervised,sun2022self,jakab2020self,wu2022casa,schmidtke2021unsupervised,jakab2020self}, multi-view data \cite{noguchi2022watch}, or point clouds \cite{you2022ukpgan}, where authors suggest to condition on the predicted keypoints and train a conditional generative model to supervise the keypoints through reconstruction losses. We propose a simpler pipeline where we apply our novel unsupervised losses to the predicted keypoints directly and do not require additional models besides the keypoint predictor itself.

\subsection{Self-supervised Learning for Point Clouds}

Self-supervised representation learning has proven to be remarkably useful in language \cite{radford2019language, devlin2018bert} and 2D vision tasks \cite{caron2019unsupervised, he2020momentum}. As LiDAR sensors become more affordable and common, there has been an increasing amount of research interest in self-supervised learning on 3D point clouds. Previous works proposed to learn representations of object or scene level point clouds through contrastive learning \cite{huang2021spatio, zhang2021self, xie2020pointcontrast} or reconstruction \cite{zhou2022self,wang2021unsupervised,yang2018foldingnet,yu2022point}, which is useful in downstream classification or segmentation tasks. In contrast, our supervision signals come from the unique structure of the human body and our learned backbone is particularly useful in downstream human keypoint estimation tasks.

%% file: sections/method.tex
\section{Method}
\label{sec:method}
\begin{figure*}
\centering
\includegraphics[width=\textwidth]{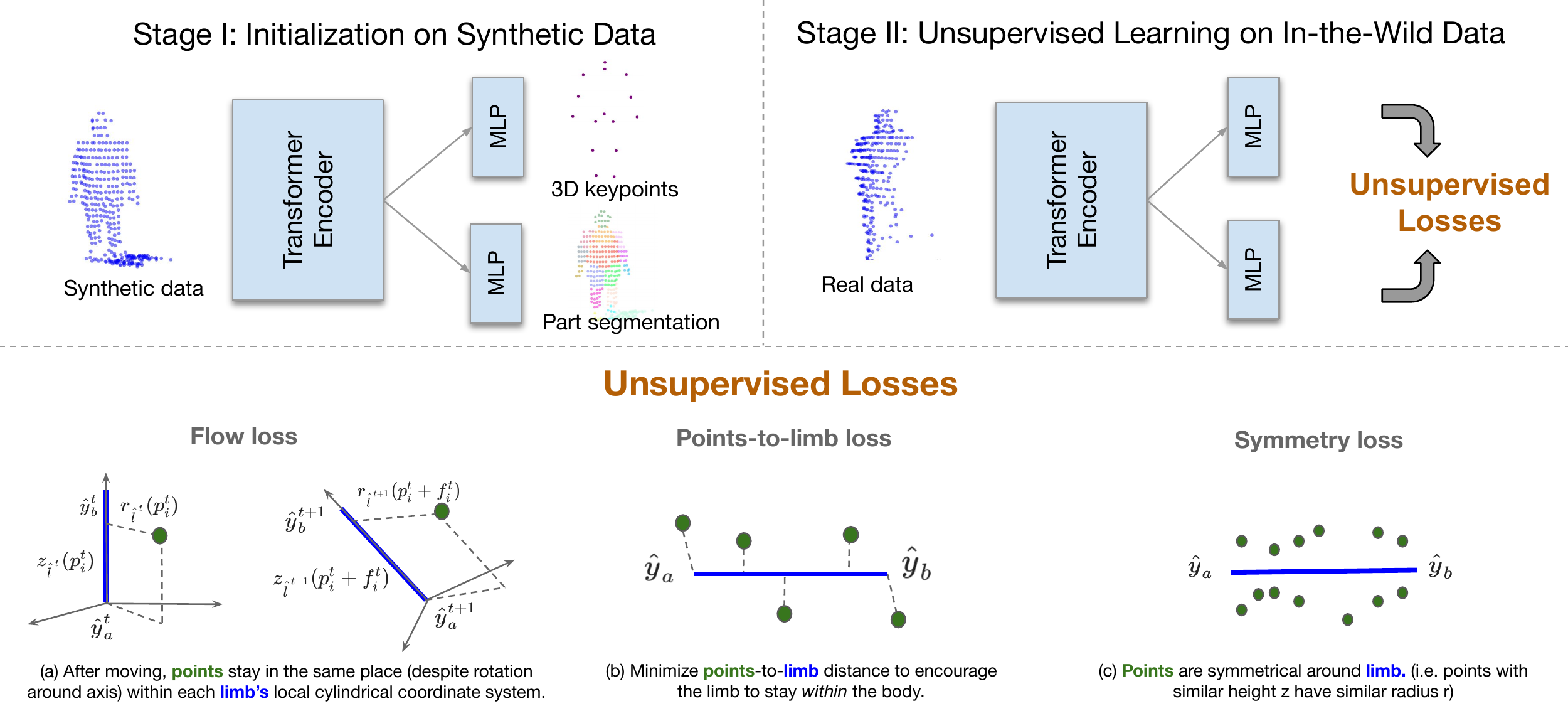}
\caption{Overview of our method. In Stage I, we warm-up the keypoint predictor and body part segmentation predictor on a small synthetic dataset. Then, in Stage II we refine the 3D keypoint predictions on a large in-the-wild dataset with unsupervised losses. The main losses are depicted on the bottom.}
\label{fig:main_figure}
\vspace{-4mm}
\end{figure*}

In this section, we describe our complete training pipeline which contains two stages. In the first stage, we initialize the model parameters on a synthetic dataset (\cref{sec:method_stage_1}). The purpose of Stage I is to warm-up the model with reasonable semantics. The second stage generalizes the model to the real-world data. In this stage, we use our unsupervised losses to refine the keypoint predictions on in-the-wild point clouds (\cref{sec:method_stage_2}). An overview of our pipeline is in \cref{fig:main_figure}.

\subsection{Stage I: Initialization on Synthetic Data}
\label{sec:method_stage_1}
In this stage, we initialize the model on a synthetic dataset that is constructed by ray casting onto randomly posed human mesh models (SMPL \cite{loper2015smpl}). We describe details of synthetic data generation in \textit{Supplementary}.

The goal of this stage is to train a model $f$ that takes a point cloud of a human $\bP \in \nR ^{N \times 3}$ and outputs 3D locations of keypoints $\hat{\bY} \in \nR^{(J+1) \times 3}$, as well as soft body part assignments (or part segmentation) $\hat{\bW} \in \nR^{N \times (J+1)}$ that contains the probability of each point $i$ belonging to body part $j \in [J]$ or the background.
\begin{gather}
\{\hat{\bY}, \hat{\bW}\} = f(\bP) \\
\forall i \in [N], \sum_{j=1}^{J+1} \hat{\bW}_{i,j} = 1
\end{gather}
Ground truth information about part segmentation $\bW$ and keypoint locations $\bY$ are readily available for synthetic data. Hence, we can train the model by directly supervising the predicted keypoint through L2 loss, 
\begin{gather}
\cL_{kp} = || \hat{\bY} - \bY ||_2
\end{gather}
and predicted segmentation through cross entropy loss,
\begin{gather}
\cL_{seg} = -\sum_{i=1}^{N} \sum_{j=1}^{J+1} \bW_{i, j} \log(\hat{\bW}_{i, j}) \label{eq:seg_loss}
\end{gather}
Overall, we minimize
\begin{gather}
\cL_{syn} = \lambda_{kp}  \cL_{\text{kp}} + \lambda_{seg} \cL_{\text{seg}} \label{eq:warmup}
\end{gather}
Notably, in \cref{subsec:ablations} we show that supervision in this stage is not required - ground truth $\bW$ and $\bY$ can be replaced by surrogate ground truths to achieve comparable results.

\subsection{Stage II: Self-Supervised Learning on In-the-Wild Data}
\label{sec:method_stage_2}

In this stage, we further refine the network using unsupervised losses. The key insight behind the design of the losses is that the human body is composed of limbs, each of which is a rigid part. Therefore, points on a limb move with the limb and should stay roughly at the same location in each limb's local coordinate system. To account for this, we propose flow loss that encourages the points to stay in the same location (despite rotation around the limb) within each limb's local cylindrical coordinate. 

We start by formally defining the key ingredients in the following formulations. In our setup, a human \textit{skeleton} $L$ is composed of \textit{limbs}, each of which is connecting two keypoints. A limb $l = (y_a, y_b) \in L$ is a line segment connecting the parent $y_a$ and child keypoint $y_b$ on this limb, and all surface points on this limb have segmentation label $a$.

All three proposed losses are in terms of surface points in each predicted limb's local coordinate system. Therefore, we first convert all input points to each limbs' local cylindrical coordinate and compute the radial and axial coordinates. Specifically, we project point $p \in \bP$ in global coordinate on to vector $\overrightarrow{\hy_a \hy_b}$, and calculate
the norm of the projected vector 
\begin{equation}
\bz(p, \hl) = \frac{(p-\hy_a) \cdot (\hy_b - \hy_a)}{||\hy_b - \hy_a||_2}
\end{equation}
and the distance between the point and $\overrightarrow{\hy_a \hy_b}$, 
\begin{equation}
\br(p, \hl) = ||p - \hy_a - \bz (\hy_b - \hy_a, \hl)||_2
\end{equation}
For simplicity, we use $\bz_{\hl}(p)$ to represent $\bz(p, \hl)$, and $\br_{\hl}(p)$ to represent $\br(p, \hl)$ in the following.

Next, we describe the formulation of each loss function in detail. 

\textbf{Flow Loss.} Flow loss considers the predictions from two consecutive frames and encourages consistency of the radial and altitude components of all points with respect to scene flow - limbs should move between frames in a way to keep radial and axial coordinates for all points constant. Formally, we define the forward and backward flow losses ($\cL_{ff}$ and $\cL_{bf}$ respectively) for limbs $\hl^t = (\hy_a^t, \hy_b^t)$ and $\hl^{t+1} = (\hy_a^{t+1}, \hy_b^{t+1})$ for predicted keypoints for timestamp $t$ and $t+1$.
\begin{multline}
\small
\cL_{ff} = \frac{1}{N}\sum_{i} \hat{\bW}_{ia}^t \cdot (|\br_{\hl^{t+1}}(p_i^{t} + f_i^{t}) - \br_{\hl^t}(p_i^{t})| + \\
 |\bz_{\hl^{t+1}}(p_i^{t} + f_i^{t}) - \bz_{\hl^t}(p_i^{t})|)
\end{multline}
\begin{multline}
\small
\cL_{bf} = \frac{1}{N} \sum_{i} \hat{\bW}_{ia}^{t+1} \cdot (|\br_{\hl^t}(p_i^{t+1} + b_i^{t+1}) - \br_{\hl^{t+1}}(p_i^{t+1})| + \\
|\bz_{\hl^t}(p_i^{t+1} + b_i^{t+1}) - \bz_{\hl^{t+1}}(p_i^{t+1})|)
\end{multline}
$f^{t}$ is the forward flow for each point $p^{t} \in \bP^{t}$ and $b^{t+1}$ is the backward flow for each point $p^{t+1} \in \bP^{t+1}$. We use Neural Scene Flow Prior \cite{li2021neural} to estimate flow for two consecutive frames of points. The overall flow loss for frame $t$ is 
\begin{align}
\small
\cL_{flow} &= \frac{1}{|L|} \sum_{\hl^t} \frac{\cL_{ff} + \cL_{bf}}{2}
\end{align}
By design, the flow loss value is the same if the radial and axial values for all points in a local coordinate system are the same in consecutive frames. This would happen if a limb in both frames are shifted in their respective orthogonal direction by the same amount. Theoretically, it is unlikely to happen for all limbs, but empirically we observe that with flow loss alone the skeleton would move out of the point cloud.
Therefore, we need additional losses to make the keypoints stay within the body. 

\textbf{Points-to-Limb Loss.}
For a predicted limb $\hl=(\hy_a, \hy_b)$, we want the points on this limb to be close to it. Hence, we introduce a points-to-limb (p2l) loss 
\vspace{-2mm}
\begin{align}
\small
    \cL_{p2l}^{\hl} &= \frac{1}{N} \sum_{i} \hat{\bW}_{ia} \bd(p_i, \hl)
\end{align}
where $\bd$ is the Euclidean distance function between a point and a line segment. We sum over all points to get the overall points-to-limb loss,
\vspace{-4mm}
\begin{align}
\small
    \cL_{\text{p2l}} &= \frac{1}{|L|} \sum_{\hl} \cL_{\text{p2l}}^{\hl}
\end{align}

\textbf{Symmetry Loss.}
Symmetry loss encourages the predicted limb $\hl$ to be in a position such that all points around this limb are roughly symmetrical around it. That is to say, points with similar axial coordinates $\bz_{\hl}$ should have similar radial values $\br_{\hl}$. To that end, we introduce symmetry loss,
\begin{align}
\small
\cL_{sym}^{\hl} &= \frac{1}{N} \sum_{i} \hat{\bW}_{ia} (\br_{\hl}(p_i) - \bar{\br}_{\hl}(p_i))^2
\end{align}
where $\bar{\br}_{\hl}(p_i)$ is the weighted mean of radial values of points with similar axial coordinates as $p_i$,
\begin{align}
\small
\bar{\br}_{\hl}(p_i) &= \frac{\sum_{j} K_h(\bz_{\hl}(p_i), \bz_{\hl}(p_j)) (\hat{\bW}_{i*} \cdot \hat{\bW}_{j*}) \br_{\hl}(p_j)} {\sum_{j} K_h(\bz_{\hl}(p_i), \bz_{\hl}(p_j)) (\hat{\bW}_{i*} \cdot \hat{\bW}_{j*})} \label{eq:sym_loss}
\end{align}
$K_h$ is Gaussian kernel with bandwith $h$, i.e. $K_h(x, y) = e^{-(\frac{x-y}{h})^2}$.
$\hat{\bW}_{i*} \in \nR^{J}$ is the $i_{th}$ row of $\hat{\bW}$, and the dot product $\hat{\bW}_{i*} \cdot \hat{\bW}_{j*}$ measures the similarity of part assignment of point $i$ and $j$, as we want the value of $\bar{r}_i^k$ to be calculated using the points from the same part as point $i$. \\
The overall symmetry loss is over all points, 
\begin{align}
\small
    \cL_{sym} &= \frac{1}{|L|} \sum_{l \in L} \cL_{sym}^{l}
\end{align}

\textbf{Joint-to-Part Loss.} In addition, we encourage each joint to be close to the center of the points on that part using a joint-to-part loss.
\begin{align}
\small
    \cL_{j2p}^{j} &= \left\lVert\hy_j - \frac{\sum_i \hat{\bW}_{ij} p_i}{\sum_i \hat{\bW}_{ij}}\right\rVert_2
\end{align}
We sum over all joints to get the overall joint-to-part loss.
\begin{align}
\small
    \cL_{j2p} &= \frac{1}{J} \sum_{j} \cL_{j2p}^{j}
\end{align}
Note that although the ground truth location of joints are not in the center of points on the corresponding part, keeping this loss is essential in making the unsupervised training more robust.

In practice, jointly optimizing $\hat{\bW}$ and $\hat{\bY}$ in Stage II leads to unstable training curves. Hence, we use the pre-trained segmentation branch from Stage I to run segmentation inference to get the segmentation labels on all of the training samples in the beginning of Stage II, and $\hat{\bW}$ is the one-hot encoding of the predicted segmentation labels.

\textbf{Segmentation Loss.} Lastly, we notice that keeping the segmentation loss at this stage further regularizes the backbone and leads to better quantitative performance. We use the inferenced segmentation $\hat{\bW}$ as the surrogate ground truth and minimize cross entropy as in \cref{eq:seg_loss}.

\textbf{Training objective.}
The overall training objective during Stage II is to minimize
\begin{multline}
\cL = \lambda_{flow} \cL_{flow} + \lambda_{\text{p2l}} \cL_{\text{p2l}} + \lambda_{sym} \cL_{sym} \\
 + \lambda_{\text{j2p}} \cL_{\text{j2p}} + \lambda_{\text{seg}} \cL_{\text{seg}}  \label{eq:overall_loss}
\end{multline}

\begin{figure}
    \centering
    \includegraphics[width=\columnwidth]{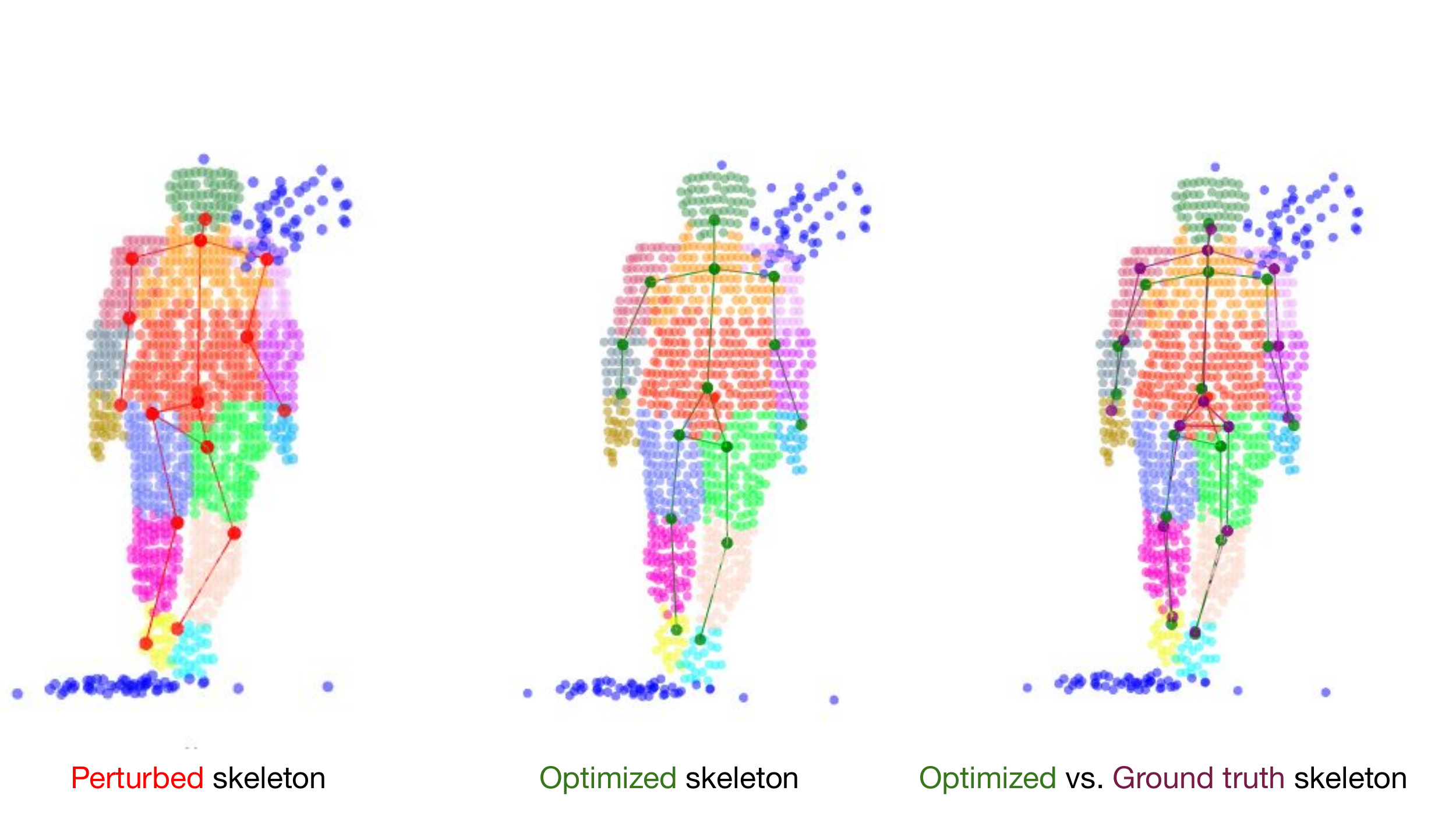}
    \caption{Effect of unsupervised losses on perturbed skeleton.}
    \label{fig:perturb_vis}
\end{figure}

To illustrate the effect of the three unsupervised losses ($\cL_{flow}$, $\cL_{p2l}$ and $\cL_{sym}$), we show the result of applying these losses on a perturbed ground truth skeleton (\cref{fig:perturb_vis}). As shown, the proposed unsupervised losses effectively moves the perturbed skeleton to locations that are closer to ground truth.

%% file: sections/experiments.tex
\section{Experiments}
\label{sec:experiments}

\subsection{Implementation Details}
The predictor model $f$ consists of a transformer backbone with fully connected layers for predicting joints and segmentation respectively. We use the same transformer backbone as in HUM3DIL \cite{zanfirhum3dil}. A fully connected layer is applied to the output of transformer head to regress the predicted $\hat{W}$ and $\hat{Y}$ respectively. There are 352,787 trainable parameters in total. We set the maximum number of input LiDAR points to 1024, and zero-pad or downsample the point clouds with fewer or more number of points. The flow is obtained using a self-supervised test-time optimization method \cite{li2021neural}. The network is trained on 4 TPUs. We train Stage I for $200$ epochs and Stage II for $75$ epochs, both with batch size 32, base learning rate of $1e-4$, and exponential decay $0.9$. Stage I and II each finishes in about 6 hours. The loss weights in \cref{eq:warmup} are $\lambda_{kp}=0.5$ and $\lambda_{seg}=1$. The loss weights in \cref{eq:overall_loss} are $\lambda_{flow}=0.02$, $\lambda_{p2l}=0.01$, $\lambda_{sym}=0.5$, $\lambda_{j2p}=2$, and $\lambda_{seg}=0.5$. The kernel bandwidth \cref{eq:sym_loss} is 0.1.

\subsection{Dataset and Metrics}
\begin{figure}[h!]
\centering
\includegraphics[width=0.9\columnwidth]{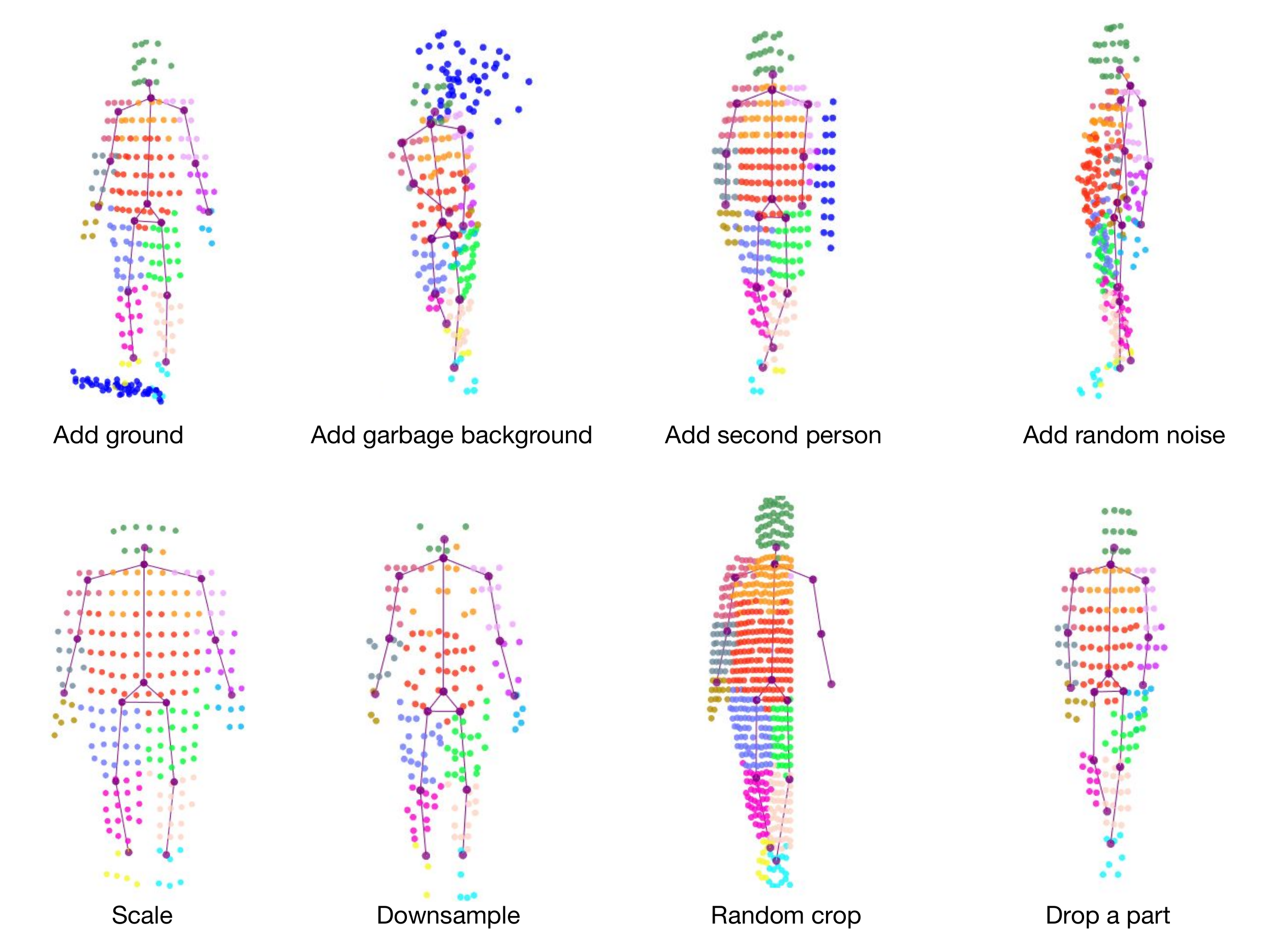}
\caption{Data augmentations applied to the synthetic point clouds (colored by ground truth segmentation labels). Ground truth skeletons are shown in \textcolor{purple}{purple}. Background points are in \textcolor{blue}{blue}.}
\vspace{4px}
\label{fig:synthetic_augmentation}
\end{figure}

We construct a synthetic dataset with 1,000 sequences of 16-frame raycasted point clouds for Stage I training. Each sequence starts with the same standing pose and ends in a random pose. We find that data augmentation is essential in Stage I training. To simulate real-world noisy background and occlusion, we apply various data augmentations to the synthetic data, including randomly downsample, random mask, add ground clusters, add background clusters, add a second person, add noise to each point, scale the person. We include examples of augmented synthetic data in \cref{fig:synthetic_augmentation}. 

In Stage II, we train on the entire Waymo Open dataset (WOD) training set (with around 200,000 unlabeled samples). As the official WOD testing subset is hidden from the public, we randomly choose 50\% of the validation set as the validation split, and the rest as the test split for benchmarking. We report average Mean Per Joint Position Error (MPJPE) on test set at the end of each stage. Formally, for a single sample, let $\hat{Y} \in \mathcal{R}^{J \times 3}$ be the predicted keypoints, $Y \in \mathcal{R}^{J \times 3}$ the ground truth keypoints, and $v \in \{0, 1\}^{J}$ the visibility indicator annotated per keypoint.
\begin{gather}
\small
\text{MPJPE} (Y, \hat{Y}) = \frac{1}{\sum_{j} v_{j}} \sum_{j \in [J]} v_j ||y_j - \hat{y}||_2
\end{gather}
Note that in this Stage,
we do Hungarian matching between the predicted and annotated keypoints per frame, and then report MPJPE on matched keypoints. We report matched MPJPE because the method is intended for scenarios where correspondence between keypoints in the unlabeled training data and downstream data is unknown.

\subsection{Results}
\begin{figure}
    \centering
    \includegraphics[width=0.8\columnwidth]{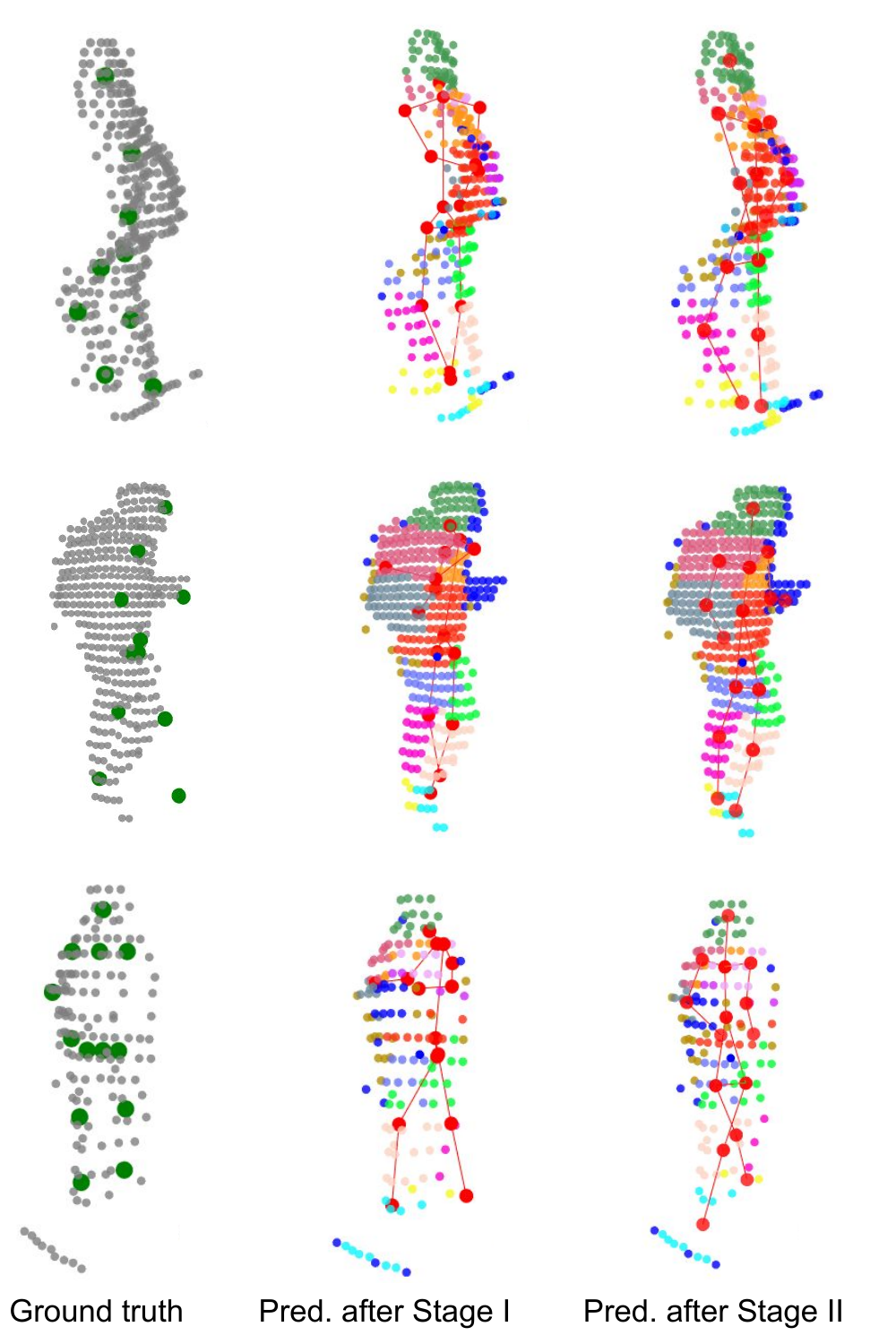}
    \caption{Visualizations of predictions on WOD at the end of Stage I and Stage II. Points are colored by predicted segmentation labels. Ground truth keypoints are in \textcolor{OliveGreen}{green} and predicted keypoints and skeletons are in \textcolor{red}{red}.}
    \label{fig:unsupervised_vis}
\vspace{-4mm}
\end{figure}

In this section we perform quantitative evaluation of \OurMethod{} at the end of Stage I and II in \cref{table:unsupervised}. Qualitative results are in \cref{fig:unsupervised_vis}. As shown, after first stage where we train on a synthetic dataset constructed from posed body models with carefully chosen data augmentations, we are able to predict reasonable human keypoints on in-the-wild point clouds. The second stage our novel unsupervised losses further refine the predicted keypoints.

\subsection{Downstream Task: Few-shot 3D Keypoint Learning}
In this experiment, we show that the backbone of our model benefits from unsupervised training on large amount of unlabeled data, and can be useful for downstream fine-tuning tasks. We start from our pre-trained backbone after Stage II, and fine-tune with annotated training samples from WOD by minimizing mean per joint error. We include few-shot experiments where we fine-tune with a extremely small amount of data  (10\% and 1\% of the training set), to represent challenging scenarios where there is a limited amount of annotated data. 

We include the LiDAR-only version of HUM3DIL (a state-of-the-art model on WOD) \cite{zanfirhum3dil} as a strong baseline. The quantitative results (\cref{table:downstream}) suggest that our backbone learns useful information from the unlabeled in-the-wild data and enables a significant performance boost on the downstream tasks. Compared to a randomly initialized backbone as used in HUM3DIL, our backbone leads to over 2 cm of decrease in MPJPE in downstream fine-tuning experiments, which is a significant improvement for the 3D human keypoint estimation task.

\begin{figure*}
    \centering
    \includegraphics[width=0.9\textwidth]{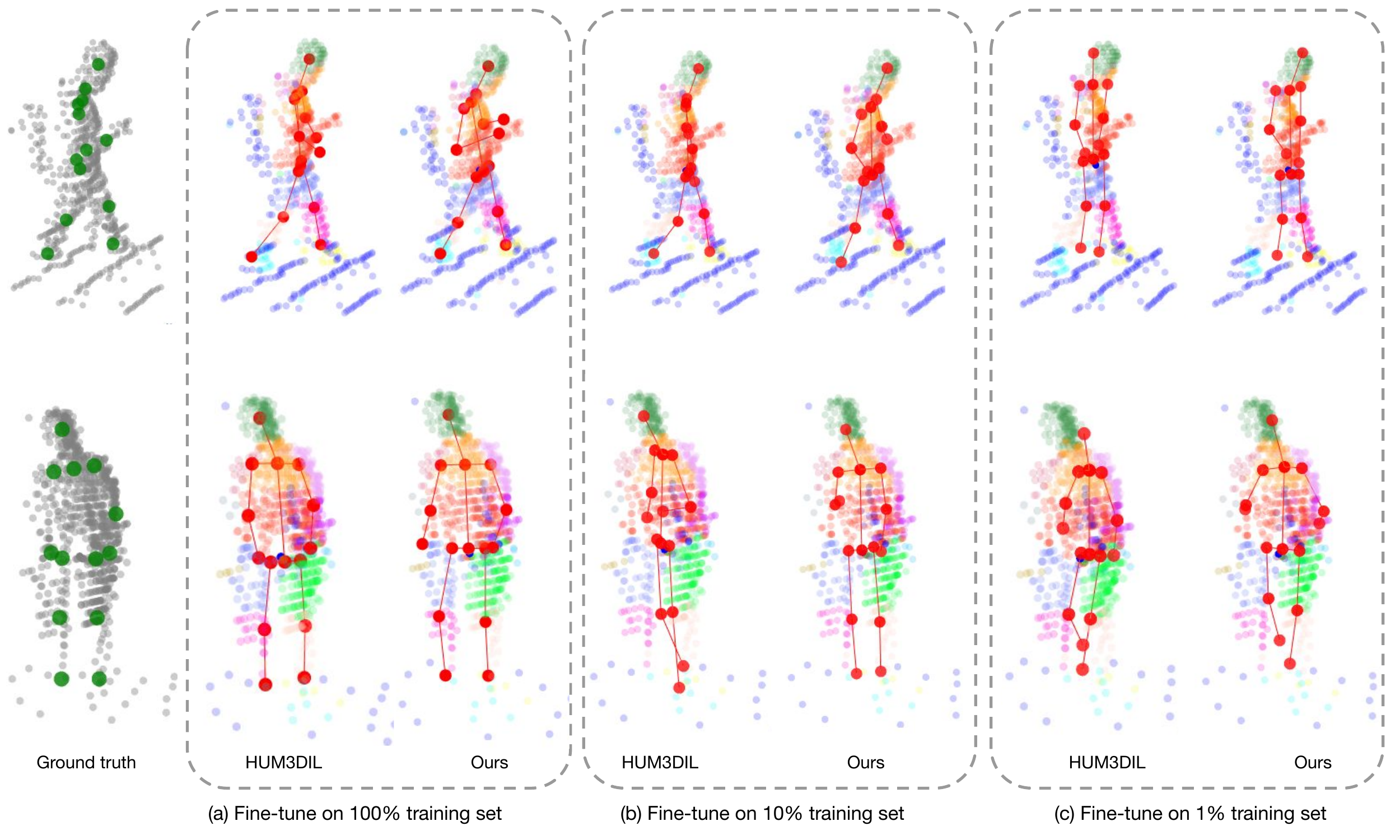}
    \caption{Predicted keypoints from fine-tuning with different amount of annotated data. The points are colored by predicted segmentation labels by our model. Predicted keypoints are shown in \textcolor{red}{red}.}
    \label{fig:downstream_vis}
\end{figure*}

We visualize the predicted keypoints under different data regime in \cref{fig:downstream_vis}. As shown, models fine-tuned from our backbone is able to capture fine details on the arms and overall produces more accurate results than HUM3DIL.

To the best of our knowledge, there does not exist previous works on completely unsupervised human keypoint estimation from point clouds. We additionally experiment with using a readout layer on top of the features learned by a state-of-the-art point cloud SSL method 3D-OAE \cite{zhang2021self}, but the MPJPE is 15 cm (compared to 10.10 cm from \OurMethod{}). Hence we consider the baselines we adopt here strong and complete. In Sec. \ref{subsec:ablations}, we further challenge our method by comparing to the domain adaptation setup and demonstrate that the performance of \OurMethod{} is still superior.

\subsection{Domain adaptation}

In the configuration where we use ground truth labels in Stage I and unsupervised training in Stage II could be seen as a domain adaption (DA) technique. Thus it is useful to compare proposed method with a commonly-used domain adaptation method. We train the same backbone model using a mix of real and synthetic data and a gradient reversal layer (aka DA loss) \cite{pmlr-v37-ganin15} to help the network to learn domain invariant keypoint features. Results in \cref{table:da} demonstrate that \OurMethod{} yields superior accuracy compared with the DA method (MPJPE 10.1 vs 11.35 cm).

\input{tables/main_tables}
\input{tables/ablations}

\input{tables/augmentations}
\subsection{Ablations}
\label{subsec:ablations}
\textbf{Effect of using GT bounding boxes in pre-processing.}
We cropped human point clouds from the entire scene by including only points within GT bounding boxes. We also conducted experiments where we train with detected bounding boxes from raw LiDAR scans using a SoTA 3D detector. Results suggest that \OurMethod{} is robust to noise in 3D detection, as there were no noticeable changes in metrics.

\textbf{Effect of synthetic dataset size.}
In our method Stage I serves as a model initialization step where we show that training on a small synthetic dataset (16,000 samples) with properly chosen data augmentations is suffice for the model to learn useful semantics. We further investigate the effect of synthetic dataset size during Stage I. We experiment with larger dataset sizes (160,000 and 1,600,000 samples) and observe that the effect of increasing synthetic dataset size is insignificant on MPJPE$_{\text{matched}}$ at the end of Stage I - it decreased from 17.7cm to 17.6cm. Lack of a notable improvements for larger dataset sizes is likely due to limited variability of generated poses in synthetic data (see Supplemental for details).

\textbf{Effect of using ground truths on synthetic data.}
While our described pipeline does not use any kind of manual labels, we do use ground truth segmentation and keypoints on synthetic dataset in Stage I because they are readily available. Here we further experiment with a variation where we do not use any kind of ground truths in Stage I (first row in \cref{table:ablations_losses}). Instead, we use KMeans clusters and cluster centers as surrogate ground truths for model initialization, similar to \cite{caron2018deep}. Note that we are able to establish correspondence between KMeans clusters from different samples due to the fact that in our data generation process, each synthetic sequence starts with the same starting standing pose. Hence, we can run KMeans clustering on the starting pose that is shared among all sequences, and for subsequent samples within each sequence, we do Hungarian matching using inter-cluster Chamfer distance to establish correspondence between clusters from consecutive frames. We observe that although initializing with surrogate ground truths leads to slightly inferior performance in Stage I, after training with the losses in Stage II the drop in performance is less visible. Overall, downstream fine-tuning performance is comparable to our best model (10.6/14.3/17.8 vs. 10.1/13.4/17.2 cm when fine-tuned on 100$\%$/10$\%$/1$\%$ of the data, see \cref{table:downstream}). This experiment suggests that method does not require any kind of ground truths, even during initialization stage.

\textbf{Effect of Losses.}
In this section we further investigate the effect of each component in our pipeline (\cref{table:ablations_losses}). First, we note that $\cL_{seg}$ in Stage I is essential because we need an initialized segmentation model to get the body part assignment for each point in order to calculate the losses in Stage II. Therefore, we only experiment with a variation of Stage I training without $\cL_{kp}$, and we observe that $\cL_{kp}$ is useful in warming up the backbone for later stages. Next, we take the backbone from Stage I (trained with both $\cL_{kp}$ and $\cL_{seg}$), and study the effect of individual losses in Stage II. Experiments No. 3/4/5 show that it is helpful to include $\cL_{j2p}$ and $\cL_{seg}$ while having all other three unsupervised losses. In experiments 6/7/8 we take out $\cL_{j2p}$ and $\cL_{seg}$, and investigate the effect of individual unsupervised losses. As shown the training becomes rather unstable if we further eliminate any of the three losses. We observe qualitatively that the metric worsens drastically because the limbs quickly move out of the human body. Experiments No. 3/4/5 suggest that $\cL_{j2p}$ and $\cL_{seg}$ are useful regularizers that make sure the limbs stay within the body, and the unsupervised losses further improve the performance by refining the keypoint location.
\vspace{-4px}
\subsection{Limitations and Future Work}
The task of keypoint location could be considered as a dual problem for semantic segmentation. In this work we use a simple segmentation network based on the same architecture as our keypoint estimation model. Using a superior segmentation model could lead to further improvements.

The proposed flow loss depends on quality of the estimated flow of LiDAR points. In this work we used a simple but reasonable method to estimate flow between two frames of LiDAR points called Neural Scene Flow prior \cite{li2021neural}. Quality of the unsupervised keypoint estimation could be improved by using a more advanced flow estimator tailored for point clouds on human body surfaces.

Lastly, we use a part of the HUM3DIL \cite{zanfirhum3dil} model which takes only LiDAR point cloud as input. The full HUM3DIL model was designed for multi-modal inputs and attains better performance. Thus, another interesting direction is to leverage multi-modal inputs.

%% file: tables/main_tables.tex
\begin{table*}
\centering
\small
\begin{tabular}{llcccc}
\toprule
Method & Backbone & \thead{Stage I\\supervised} & \thead{1\% training set\\MPJPE cm. \color{OliveGreen}{(gain)}}  &  \thead{10\% training set\\MPJPE cm. \color{OliveGreen}{(gain)}}  & \thead{100\% training set\\MPJPE cm. \color{OliveGreen}{(gain)}}  \\
\midrule
HUM3DIL \cite{zanfirhum3dil} & Randomly initialized &  & 19.57 & 16.36 & 12.21 \\
\multirow{4}{*}{\OurMethod{}} & Pre-trained on synthetic only & \ding{52} & 18.52 \color{OliveGreen}{(-1.05)}  &  15.10  \color{OliveGreen}{(-1.26)}  & 11.27 \color{OliveGreen}{(-0.94)} \\
 & Pre-trained on 5,000 WOD-train & \ding{52} & 17.87 \color{OliveGreen}{(-1.70)}  & 14.51 \color{OliveGreen}{(-1.85)}   &10.73  \color{OliveGreen}{(-1.48)} \\
 & Pre-trained on 200,000 WOD-train &  & 17.80 \color{OliveGreen}{(-1.77)}  &  14.30 \color{OliveGreen}{(-2.06)}  &  10.60 \color{OliveGreen}{(-1.61)}  \\
 & Pre-trained on 200,000 WOD-train & \ding{52} & \textbf{17.20} \color{OliveGreen}{(\textbf{-2.37})}  &  \textbf{13.40} \color{OliveGreen}{(\textbf{-2.96})}  &  \textbf{10.10} \color{OliveGreen}{(\textbf{-2.11})}  \\
\bottomrule
\end{tabular}%

\caption{Downstream fine-tuning results. Check marks in ``Stage I supervised" mean that we use ground truth part labels in Stage I, otherwise we use KMeans labels. }
\label{table:downstream}
\vspace{-4mm}
\end{table*}

\begin{table}
\centering
\begin{tabular}{lc}
\toprule
Training data & {MPJPE\textsubscript{matched}} ($\downarrow$)  \\
\midrule
Synthetic only &  17.70    \\
5,000 WOD-train & 14.64   \\
200,000 WOD-train & 13.92   \\
\bottomrule
\end{tabular}%
\vspace{-4px}
\caption{Unsupervised learning (Stage II) results.}
\label{table:unsupervised}
\end{table}

\begin{table}
\centering
\begin{tabular}{lcc}
\toprule
Domain distribution & DA loss & {MPJPE} ($\downarrow$)  \\
\midrule
100\% real  & & 12.21 \\
50/50\% real/synthetic &  & 12.08   \\
50/50\% real/synthetic & \ding{52} & 11.35   \\
\bottomrule
\end{tabular}%
\caption{Unsupervised domain adaptation results evaluated on WOD validation set.}
\label{table:da}
\vspace{-4mm}
\end{table}

%% file: tables/ablations.tex
\begin{table*}[h!]
\centering
\resizebox{\textwidth}{!}{\begin{tabular}{@{}cc|ccc|lllllc@{}}
\toprule
 & & \multicolumn{3}{c}{\large Stage I} & \multicolumn{6}{|c}{\large Stage II}  \\ \midrule
No. &  \multicolumn{1}{c|}{Exp.} & $\mathcal{L}_{kp}$  &  $\mathcal{L}_{seg}$   & MPJPE$_{\text{matched}}$ &  $\mathcal{L}_{j2p}$  &  $\mathcal{L}_{seg}$  &  $\mathcal{L}_{sym}$  & $\mathcal{L}_{p2l}$  &  $\mathcal{L}_{flow}$ & MPJPE$_{\text{matched}}$ \\ \midrule \midrule
1 &\multirow{1}{7cm}{\centering Effect of using KMeans labels in Stage I}&  \ding{52} & \ding{52} & 19.2 & \ding{52} & \ding{52} & \ding{52} & \ding{52} & \ding{52} & 14.5 \\ \midrule \midrule
2&\multirow{1}{7cm}{\centering Effect of $\mathcal{L}_{kp}$ in Stage I}  &  &  \ding{52}    & N/A & \ding{52} & \ding{52} & \ding{52} & \ding{52} & \ding{52} & 14.2 \\ \midrule \midrule
3
 &\multirow{3}{7cm}{\centering Effect of warmup losses in Stage II} &  & & & & \ding{52} & \ding{52} & \ding{52} & \ding{52} &  15.0 \\
4& &  & & & \ding{52} & & \ding{52} & \ding{52} & \ding{52} &  14.2   \\
5& &  & & & & & \ding{52} & \ding{52} & \ding{52} &  15.2   \\ \midrule
6 & \multirow{8}{7cm}{\centering Effect of unsupervised losses in Stage II}  & & & &  &  & & \ding{52} & \ding{52} & 30.1  \\
7 & & & & &  &  & \ding{52} & & \ding{52} & 15.6  \\
8 & & & & &  &  & \ding{52}& \ding{52} & & 25.7  \\
9 & & & & & \ding{52} & \ding{52} & & \ding{52} & \ding{52} & 14.3 \\
10 & & & & & \ding{52} & \ding{52} & \ding{52} & & \ding{52} & 14.9  \\
11 & & & & & \ding{52} & \ding{52} & \ding{52} & \ding{52} & & 14.4 \\
12 & & & & & \ding{52} & \ding{52} & & &  & 14.9  \\ \midrule\midrule
Full model (\OurMethod{}) & & \ding{52}    &  \ding{52}    &  \textbf{17.7} & \ding{52} & \ding{52} & \ding{52} & \ding{52} & \ding{52} & \textbf{13.9}  \\ 
\bottomrule
\end{tabular}}
\caption{Ablations studies on the effect of individual loss term in our method. Experiments 3 through 12 are using both losses in Stage I. Full model is using GT labels for Stage I.}
\label{table:ablations_losses}
\vspace{-4mm}
\end{table*}

%% file: sections/conclusion.tex
\vspace{-4px}
\section{Conclusion}
\label{sec:conclusion}
In this work, we approached the problem of 3D human pose estimation using points clouds in-the-wild, introduced a method (\OurMethod{}) for learning 3D human keypoints from point clouds without using any manual 3D keypoint annotations. We shown that the proposed novel losses are effective for unsupervised keypoint learning on Waymo Open Dataset. Through downstream experiments we demonstrated that \OurMethod{} can additionally serve as a self-supervised representation method to learn from large quantity of in-the-wild human point clouds. In addition, \OurMethod{} compares favorably with a commonly used domain adaptation technique. The few-shot experiments empirically verified that using only 10$\%$ of available 3D keypoint annotation the fine-tuned model reached comparable performance to the state-of-the-art model training on the entire dataset. These results opens up exciting possibility to utilize massive amount of sensor data in autonomous driving to improve pedestrian 3D keypoint estimation.